# Geospatial foundation models for image analysis: evaluating and enhancing NASA-IBM Prithvi's domain adaptability

Chia-Yu Hsu[1], Wenwen Li[1] *, Sizhe Wang[1,2]

[1] School of Geographical Sciences and Urban Planning, Arizona State University, Tempe, AZ 85287
[2] School of Computing and Augmented Intelligence, Arizona State University, Tempe, AZ 85287

* Correspondence: wenwen@asu.edu

Abstract: Research on geospatial foundation models (GFMs) has become a trending topic in geospatial artificial intelligence (AI) research due to their potential for achieving high generalizability and domain adaptability, reducing model training costs for individual researchers. Unlike large language models, such as ChatGPT, constructing visual foundation models for image analysis, particularly in remote sensing, encountered significant challenges such as formulating diverse vision tasks into a general problem framework. This paper evaluates the recently released NASA-IBM GFM Prithvi for its predictive performance on high-level image analysis tasks across multiple benchmark datasets. Prithvi was selected because it is one of the first open-source GFMs trained on time-series of high-resolution remote sensing imagery. A series of experiments were designed to assess Prithvi's performance as compared to other pre-trained task-specific AI models in geospatial image analysis. New strategies, including band adaptation, multi-scale feature generation, and fine-tuning techniques, are introduced and integrated into an image analysis pipeline to enhance Prithvi's domain adaptation capability and improve model performance. In-depth analyses reveal Prithvi's strengths and weaknesses, offering insights for both improving Prithvi and developing future visual foundation models for geospatial tasks.

1. Introduction

Geospatial Artificial Intelligence (GeoAI) is an interdisciplinary field that integrates geospatial data science with artificial intelligence techniques to solve complex spatial problems (Janowicz *et al.* 2020, Li 2020). One of its most promising applications is in the realm of image analysis, particularly in interpreting and extracting valuable information from remote sensing imagery (Li, Wang, *et al.* 2022, Udawalpola *et al.* 2022). Remote sensing imagery has revolutionized the way we understand the physical process on the Earth's surface and in the atmosphere. Utilizing sensors mounted on satellites, drones, or aircraft, remote sensing captures high-resolution images. These images provide invaluable data across various sectors, including environmental monitoring (VoPham *et al.* 2018), disaster management (Mahmood 2022), agriculture (García Pereira *et al.* 2020), and urban planning (Alastal and Shaqfa 2022), among others. Traditional methods of image analysis, such as thresholding techniques, often require manual intervention and are time-consuming (Zhou *et al.* 2019). This makes them less efficient for handling large datasets. GeoAI, on the other hand, leverages machine learning algorithms to automatically analyze and interpret geospatial images, thereby significantly improving the speed and accuracy of data extraction (Li and Hsu 2022). As GeoAI continues to evolve, it opens up new avenues for more automated and intelligent image analysis, making it a subject of keen interest for researchers and practitioners in the geospatial domains (Li and Hsu 2020, Gao *et al.* 2023).

Recent advancement in deep learning have significantly improved the capabilities of geospatial image analysis, but these models come with their own set of challenges. One pressing issue is the requirement for large, annotated datasets for effective training (Deng *et al.* 2009, Lin *et al.* 2014). This is particularly challenging in specialized fields such as remote sensing, where the need for domain-specific knowledge, the large volume of data, and data variability due to factors such as seasonal changes make obtaining annotated data both time-consuming and expensive (Li *et al.* 2021). Furthermore, these models are generally task-specific, meaning that a model trained for one application may not easily generalize to another without substantial retraining and fine-tuning (Zhang *et al.* 2020, 2021). The computational cost is another significant barrier, as deep learning models often require specialized hardware for both training and inference (Shankar and Reuther 2022).

The recent advances in AI foundation models present a compelling solution to some of these limitations. Unlike deep learning models that require large training datasets for each specific task, foundation models are often pre-trained using self-supervised learning (SSL) on vast amounts of data (Zhou *et al.* 2023). This allows researchers to fine-tune these models on relatively smaller, task-specific annotated datasets, thereby reducing the annotation burden (Gu *et al.* 2023, Wang, Feng, *et al.* 2023, Li *et al.* 2024). Additionally, foundation models are designed to generalize across a variety of data analysis tasks, eliminating the need for separate models and the associated retraining for each specific application (OpenAI 2023, Touvron *et al.* 2023). While it is true that foundation models also require substantial computational resources for initial training, their ability to generalize across tasks means that the computational cost can be amortized over multiple applications, making them a more efficient choice for organizations that require solutions for a range of geospatial analysis tasks, as well as smaller research groups which do not have expertise or computational resources for building such large models.

Prithvi, a geospatial foundation model developed by NASA and IBM, sets itself apart from other AI foundation models in vision tasks through several distinctive features (Jakubik *et al.* 2023). Pre-trained on contiguous US Harmonized Landsat Sentinel 2 (HLS) data, Prithvi is uniquely equipped to process remote sensing images in time series. This capability is often absent in other foundation models and enables Prithvi to well perform in various downstream tasks, such as burn scars segmentation, flood segmentation, and land cover classification. Additionally, Prithvi is designed to work with a 6-band input, including Red (R), Green (G), Blue (B), Narrow Near Infrared (NIR), Short Wave Infrared (SWIR) 1, and SWIR 2, as opposed to the conventional RGB imagery used in most AI foundation models. This multi-band capability enhances Prithvi's ability to capture a wider range of spectral information, thereby increasing its versatility and applicability across a diverse set of geospatial data and tasks.

Despite advancement in GeoAI and foundation models, there is a notable gap in the literature concerning the performance evaluation of geospatial foundation models such as IBM's Prithvi in the realm of remote sensing image analysis, especially in environmental feature detection and segmentation. Unlike general-purpose AI foundation models, Prithvi is pre-trained on remote sensing images. This unique training dataset raises a presumption that Prithvi may offer inherent advantages in geospatial tasks over other pre-trained models. In addition, Prithvi's unique 6-band input capability, as opposed to the conventional RGB imagery, could have significant implications for its applicability and performance in real-world geospatial applications. To substantiate these presumptions, we raised the following research question: "How does IBM's Prithvi perform in geospatial image analysis tasks as compared to other pre-trained task-specific AI models?"

To answer this question, the study's objectives include a detailed performance evaluation of Prithvi on challenging image analysis tasks, such as object detection and instance segmentation. Four remote sensing datasets covering diverse geographical regions and features—including a natural feature dataset, a global Mars crater dataset, an Arctic permafrost landscape dataset, and an agricultural land dataset (EuroCrops from central Europe)—are used in the analysis. Since the Prithvi model primarily provides a feature extraction backbone (the encoder part), several new strategies were introduced to adapt it to downstream tasks. These include band adaptation, a multi-scale decoder, and a new fine-tuning strategy designed to maximize its predictive performance.

The remainder of the paper is structured as follows. Section 2 reviews research on GeoAI and recent development of foundation models for geospatial image analysis. Section 3 introduces the four datasets used in this work, providing an overview of their characteristics and geographical distribution. Section 4 details the adaptations and enhancements applied to the Prithvi model, outlining the methodology and analyses conducted to evaluate Prithvi's performance. Section 5 presents the evaluation and experimental results, followed by Section 6, which discusses the strengths and weaknesses of the Prithvi model. Section 7 concludes the paper with a summary of key insights and suggestions for future research.

2. Literature review
    2.1. GeoAI and geospatial image analysis

GeoAI is an interdisciplinary field that combines the predictive power of AI with the intricacies of geospatial data science, offering a unique approach to solving complex spatial problems (VoPham *et al.* 2018, PS Chauhan and Shekhar 2021). Within this context, the application of GeoAI in image analysis stands out for its unique challenges and opportunities. One key challenge is the complexity of the data involved. Unlike traditional image analysis, GeoAI deals with multispectral and multi-band images, often captured through advanced remote sensing technologies (Li and Hsu 2022). These datasets are not only large in scale but also diverse in nature, incorporating multiple sources such as satellites, aerial photographs, and ground-based sensors (Wang and Li 2021). This data heterogeneity, coupled with the temporal dynamics inherent in geophysical phenomena such as hurricanes and wildfire, requires spatially and spatiotemporally explicit algorithms capable of interpreting intricate spatial and temporal relationships. Meanwhile, GeoAI must account for various uncertainties, such as sensor errors and missing data, making the algorithms robust and adaptable. Despite these complexities, GeoAI has proven to be an indispensable tool in a range of applications, from environmental monitoring to urban planning (VoPham *et al.* 2018, Kamel Boulos *et al.* 2019, Liu and Biljecki 2022). Its ability to handle these multifaceted challenges sets it apart from traditional methods and makes it a focal point of contemporary geospatial research.

The integration of deep learning into GeoAI has significantly advanced the field of image analysis. Initially, the direct application of deep learning models to geospatial tasks yielded mixed results, often due to the complexities inherent in geospatial data (Lee 2019, Bhuiyan *et al.* 2020, Li and Hsu 2020). To overcome these limitations, the field has evolved to incorporate expert knowledge, thereby enhancing the models' interpretability and effectiveness in handling the unique challenges of geospatial data (Janowicz *et al.* 2020, Hsu *et al.* 2021, Li *et al.* 2021). This integration of domain expertise has been a pivotal advancement, allowing for more reliable solutions in GeoAI applications. Separately, the field has also embraced transfer learning, particularly useful in scenarios where acquiring extensive labeled datasets is time-consuming and costly. This approach allows researchers to fine-tune pre-trained models for specific geospatial tasks. Alongside these advances, there has been a growing focus on models that understand spatial hierarchies, crucial for complex tasks such as urban planning (Stubbings *et al.* 2019, Zhou *et al.* 2021). However, challenges remain in creating models that can generalize across multiple tasks without extensive model retraining. This sets the stage for the emergence of foundation models, which offer a more unified and adaptable framework for handling the complex and diverse nature of geospatial data.

    2.2. Visual foundation models

Foundation models have emerged as a transformative force in computer vision, as they offer a robust framework and an appealing prospect to facilitate domain adaptation with low computational cost. Several innovative elements contribute to their transformative impact. First, foundation models process extensive and diverse datasets, setting them apart from traditional models that often operate on small and domain-specific data (Bommasani *et al.* 2021, Li, Xu, *et al.* 2023). This broad data scope is crucial for capturing representative image features and patterns, and it sets the stage for foundation models' second defining characteristic: pre-training methodologies. Due to the sheer volume of data, traditional supervised learning approaches are often impractical, leading to a new strategy of self-supervised learning (SSL) for pre-training the foundation models (Awais *et al.* 2023). Because of this, the models are capable to generalize across a multitude of tasks (Yuan *et al.* 2021, OpenAI 2023, Touvron *et al.* 2023). Third, foundation models also allow for fine-tuning from a domain-specific dataset to further boost its performance and domain adaptation (Zhou *et al.* 2023). Collectively, these defining characteristics make foundation models both revolutionary and complex tools in the image analysis landscape.

In the realm of computer vision and image analysis, foundation models have been categorized into various types based on their prompting mechanisms and data modalities, as outlined by Awais et al. (2023). Textually prompted models like CLIP (Contrastive Language-Image Pre-training; Radford *et al.* 2021) and ALIGN (A Large-scale ImaGe and Noisy-Text Embedding; Jia *et al.* 2021) interpret visual data through text-based prompts, leveraging extensive image-text datasets for pre-training and visual question answering. Visually prompted models such as Segment Anything Model (SAM; Kirillov *et al.* 2023) and SegGPT (Wang, Zhang, Cao, *et al.* 2023) utilize visual cues such as bounding boxes or segmentation masks and often rely on partially synthetic datasets with pseudo labels. Heterogeneous modality models like CLIP2Video (Fang *et al.* 2021) and AudioCLIP (Guzhov *et al.* 2022) integrate multiple types of data—vision, text, and audio—for a more comprehensive understanding of the visual world. Lastly, generalist models like VisionLLM (Wang, Chen, *et al.* 2023) exemplify the ability to generalize across a multitude of tasks when provided with appropriate prompts. These models not only embody the defining characteristics of foundation models but also showcase the adaptability and diversity that make them a cornerstone in modern computer vision research.

While foundation models have made significant strides in general-purpose computer vision, they come with their own set of limitations (Awais *et al.* 2023). One key issue is their limited contextual understanding, which can lead to a lack of depth when tackling geospatial tasks. For example, a general-purpose model might excel at a wide array of object recognition tasks but may struggle with the semantic interpretations required in specialized scientific or industrial applications (Kirillov *et al.* 2023). In addition, the use of diverse training data without human assessment can introduce biases or inaccuracies, leading to a propagation of such errors in downstream analyses (Glocker *et al.* 2022, Wójcik 2022). This issue is further compounded by the difficulty in customizing these models to achieve expert-level performance in a specific scientific domain. These limitations have led to a growing interest in specialized foundational models that are trained on data pertinent to a research field (Alfassy *et al.* 2022, Nguyen *et al.* 2023, Tu *et al.* 2023, Wu *et al.* 2023). These models aim to marry the generalizability and adaptability of foundation models with the knowledge required for domain-specific tasks.

### 2.3. Geospatial foundation models

The quest for precision and contextual sensitivity in specific scientific domains has catalyzed the development of specialized foundation models. In the realm of geospatial analysis, this pursuit has led to the emergence of Geospatial Foundation Models (GFMs; Mai *et al.* 2023). Unlike their general-purpose counterparts, GFMs are designed to interpret the complex patterns of the Earth's surface and atmosphere. They address challenges such as spatial heterogeneity (Sun *et al.* 2023), temporal dynamics (Yao *et al.* 2023) and the multidimensional nature of geospatial data (Jakubik *et al.* 2023), marking an advancement in how we analyze and understand our planet.

In developing GFMs, Transformers have emerged as the preferred architecture, attributed to their superior management of long-range dependencies and the implementation of a dynamic attention mechanism. This allows Transformers to focus selectively on image segments, emphasizing features crucial for specific tasks. Notably, the Vision Transformer (ViT; Dosovitskiy *et al.* 2021) revolutionized image analysis by treating images as sequences of patches, leading to its widespread adoption (Cha *et al.* 2023, Sun *et al.* 2023, Wang, Zhang, Xu, *et al.* 2023, Dimitrovski *et al.* 2024). Building on this, the Swin Transformer (Liu *et al.* 2021) introduces a hierarchical design that enhances image processing efficiency (Sun *et al.* 2023). However, addressing complex challenges such as multiscale issues in spatial and temporal dimensions requires further enhancements to these models. For example, Jakubik *et al.* (2023) improved temporal data handling by integrating temporal information as channels. Yao *et al.* (2023) developed a three-branch network utilizing the Video Swin Transformer (Liu *et al.* 2022) to harmonize spatial affinity, temporal continuity, and spatiotemporal interaction. In addition, the approach of refining established foundation models for specific geospatial tasks illustrates how techniques from conventional image analysis can be adapted to meet the unique demands of remote sensing and geospatial applications. For instance, SAM (Kirillov *et al.* 2023) excels in object segmentation within images without predicting category information. Yan *et al.* (2023) enhanced SAM's segmentation capabilities for category-specific tasks by incorporating a new mask decoder and introducing a prompt encoder designed for SAR imagery, leveraging SAR-specific prompts. Similarly, Chen, Liu, *et al.* (2024) leveraged SAM for instance segmentation in remote sensing images, augmenting the model with a novel prompt learning technique. These adaptations showcase the potential of Transformers in addressing the complex needs of remote sensing imagery analysis and geospatial applications, demonstrating their versatility and effectiveness across a broad range of geospatial contexts.

Training GFMs predominantly utilizes Masked Autoencoders (MAE; He *et al.* 2022) due to their effectiveness in self-supervised learning (SSL) methodologies for imagery, offering a scalable approach to training without the need for labeled data. This SSL method, by obscuring parts of the input images and learning to reconstruct them, enables the models to learn rich representations of geospatial features and dynamics autonomously. However, certain scenarios necessitate supervised training and fine-tuning, particularly for downstream tasks or when models are developed with specific functionalities in mind, like segmentation that requires category information (Yan *et al.* 2023, Yao *et al.* 2023).

In our examination of GFMs, IBM's Prithvi (Jakubik *et al.* 2023) stands out for its unique approach to GeoAI and geospatial data analysis, prompting us to select it for detailed evaluation. Prithvi is unique among AI foundation models for its design that accommodates a 6-band input, including Red, Green, Blue, NIR, SWIR 1, and SWIR 2. This capability allows it to capture a broader spectrum of spectral information than the conventional RGB imagery, enhancing its versatility and efficacy across various geospatial tasks. In addition to its advanced spectral analysis capabilities, Prithvi has the advantage in its scalability through processing a large dataset across the continental US. Furthermore, as an open-source model, Prithvi encourages wider access and community-driven enhancements. It includes the release of trained model weights, allowing researchers to directly fine-tune it for a variety of downstream tasks, thus amplifying its utility and applicability in real-world scenarios. Beyond prior evaluations focusing on its semantic segmentation capabilities in flood mapping (Li, Lee, *et al.* 2023), our assessment of Prithvi extends to its domain adaptability in other crucial image analysis tasks using multiple datasets. Additionally, several enhancement strategies are applied on top of the native Prithvi model to maximize its potential for such analyses.

3. Data

In our evaluation on the Prithvi model, we utilized a total of four datasets, tailored to support two distinct visual recognition tasks: object detection and instance segmentation. Table 1 summarizes these datasets, including the number of input image bands, image sizes, number of training and testing images, and number of object classes. In the subsequent sections, we delve deeper into the details of each dataset:

**Mars crater dataset**. The Mars crater dataset, employed in the 2022 GeoAI Martian Challenge, represents a comprehensive and varied collection of 102,675 images sourced from a global mosaic of Mars. Constructed using Mars Odyssey's Thermal Emission Imaging System (THEMIS) daytime infrared (DIR) data, the mosaic delivers a 100 m resolution covering Mars's entire surface, as documented by Edwards *et al.* (2011). Each image captures a 25.6 km by 25.6 km area, presented in 256 × 256 pixels, offering a detailed and representative snapshot of Martian terrain. Over 301,912 craters are annotated by Geology experts with instance-level bounding boxes, drawing on the extensive Mars impact crater catalog by Robbins and Hynek (2012), a compilation from multiple rounds of manual reviews of infrared imagery and topographic data, documenting over 640,000 craters with detailed positional, morphological, and morphometric information. Following the process described by Hsu *et al.* (2021), the dataset was developed by extracting non-overlapping samples from the global mosaic, applying distortion correction, and addressing partially visible craters. The dataset showcases diverse crater sizes, from as small as 0.7 km to as large as 25.5 km in diameter. This diversity poses a unique challenge, requiring models to accurately discern features in both sparsely and densely cratered landscapes.

Table 1. Benchmark datasets used for evaluating the Prithvi model.

| Evaluation task | Object detection | | Instance segmentation | |
|---|---|---|---|---|
| Dataset Statistics | Mars crater | Earth's natural feature | Ice-wedge polygon | EuroCrops |
| Dataset split (training/testing) | 9000/1000 | 575/251 | 735/132 | 755/189 |
| Bands | 3 | 3 | 3 | 6 |
| Class count | 1 | 8 | 1 | 5 |
| Image size (Min, Max, and Median) Unit: pixel | 256 × 256 | 217 × 232 1000 × 1000 540 × 350 | 199 × 199 507 × 507 203 × 203 | 128 × 128 |
| Statistics on objects count per image (Min, Max, and Median) | 1, 27, 3 | 1, 7, 1 | 1, 447, 27 | 1, 53, 24 |
| Statistics on object size (Min, Max, Median) Unit: pixel | 8 × 7 255 × 255 75 × 195 | 40 × 50 994 × 1000 324 × 277 | 1 × 1 506 × 504 143 × 100 | 2 × 3 128 × 128 58 × 79 |

To tailor the dataset for our experiments, specifically aiming for tests on unseen areas, we resampled the dataset, focusing on images from latitudes between 30 and -30 degrees. The training set was adjusted to include 9,000 images, and the testing set now consists of 1,000 images. Despite this modification, we ensured the geographical distribution of images within each set remains consistent with the original dataset's broad coverage. This approach guarantees that the training and testing sets accurately reflect a balanced representation of Martian terrain, supporting a thorough and equitable evaluation of model effectiveness.

**Earth's natural feature dataset**. The development of the natural feature dataset, as reported in the work by Li and Hsu (2020), represents a significant effort to compile a diverse collection of environmental features crucial for advancing research in geospatial analysis and landscape scene understanding. This foundational dataset for the current study includes over 100 manually labeled remote-sensing images for each of eight distinct natural features: craters, volcanoes, rivers (encompassing both meandering and non-meandering types), lakes, sand dunes, hills, and iceberg tongues. The initial phase involved utilizing geographical gazetteers, with an emphasis on the United States Geological Survey (USGS) Geographic Name Information System (GNIS), for accurate identification and categorization of various terrain objects.

This was followed by gathering and labeling images from multiple sources such as Google Earth, the USGS Earth Explorer, and relevant Google Images search results.

This dataset consists of a moderate number of images with a relatively low density of objects per image. The features in this dataset exhibit significant size variability, from medium sized to very large (see statistics in Table 1), reflecting the natural contours and diverse scales of these features. The varying image sizes add to the complexity of the detection task. The imagery has diverse spatial resolutions and spectral bands, including 1-meter optical imagery from the USGS National Agriculture Imagery Program, as well as sub-meter optical images and 2-meter multi-spectral images from DigitalGlobe's Worldview-2 satellite. Each image in this extensive dataset is accompanied by detailed annotation data, including bounding boxes to accurately delineate the terrain features of interest. The compilation of this dataset not only facilitates the current study but also supports a wide range of research avenues, particularly in the development of landscape scene recognition techniques.

**Ice-wedge polygon dataset**. The Ice-wedge polygon (IWP; Bhuiyan *et al.* 2020) dataset stands as a critical resource for mapping permafrost landscapes at a pan-Arctic scape. This dataset contains a collection of 867 image tiles with 34,931 annotated IWPs spread across a diverse array of tundra vegetation types, including sedge, tussock, and barren tundra (Li, Hsu, *et al.* 2022). This dataset is distinguished by its precision in annotation, featuring instance segmentation masks that accurately delineate each IWP, thereby facilitating fine-grained image analysis tasks. Originating from very high-resolution (0.5m) imagery captured by Maxar sensors, the dataset highlights the variability and complexity of the tundra landscape. The dataset has a higher density of feature distribution within the image scenes compared to other datasets (see statistics in Table 1). The sizes of these features vary significantly, ranging from one pixel to nearly covering the entire image scene.

**EuroCrops**. The EuroCrops dataset is the most comprehensive open-access dataset in the European Union, featuring 944 image scenes and corresponding crop land labels captured in April 2019 (Schneider *et al.* 2021). The dataset, derived from two cloud-free Top of Atmosphere (TOA) Sentinel-2 images, offers a spatial resolution of 10 meters, focusing on central Denmark's agriculturally rich and flat terrains. Each image scene is sized at 128 by 128 pixels, with detailed labels on five cropland classes: spring cereal, winter cereal, maize, grassland, and "other". The features vary from very small (2 by 3 pixels) to quite large (128 by 128 pixels) within the fixed-size image scenes (128 by 128 pixels). This dataset has a moderate to high density of objects per image, with a wide range of object sizes (see statistics in Table 1). The 6-band nature of this dataset makes it particularly helpful for evaluating the advantages of Prithvi, which is also trained on six-band remote sensing imagery, to support agriculture research. Figure 1 demonstrates a few samples from each benchmark dataset.

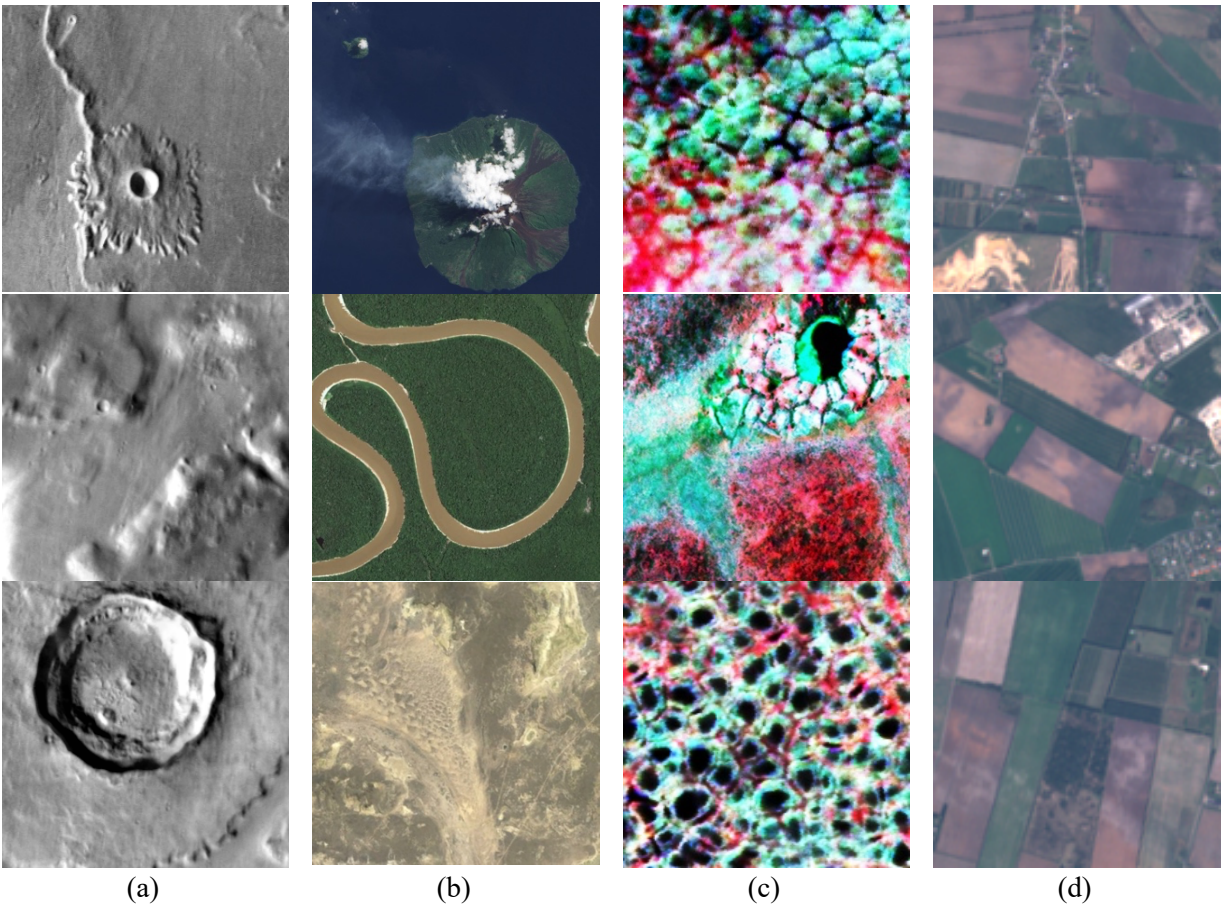

(a) (b) (c) (d)

Figure 1. Sample images from the four benchmark datasets. (a) Mars crater. (b) Earth's natural feature. (c) Ice-wedge polygon. (d) EuroCrops.

4. The Prithvi model, task-specific adaptation, and model enhancement

4.1. Model architecture and pre-training

In the development of NASA-IBM's Prithvi model, the pre-training phase plays a crucial role (Jakubik *et al.* 2023). The model is trained on HLS data, a dataset that fuses measurements from multiple satellite sensors, including NASA/USGS Landsat 8 and 9's Operational Land Imager (OLI) and Europe's Copernicus Sentinel-2A and Sentinel-2B's Multi-Spectral Instrument (MSI; Masek *et al.* 2021). To ensure the consistency and reliability of this data, the HLS project employs algorithms for atmospheric correction, cloud and cloud-shadow masking, and spatial co-registration. Specifically, Prithvi was trained on the HLSL30 product, which offers a 30-meter spatial resolution and is provided in a Cloud Optimized GeoTIFF (COG) format. The training data spanned the continental United States for the year 2017 and focused on six spectral bands, namely Blue, Green, Red, Narrow NIR, SWIR 1, and SWIR 2.

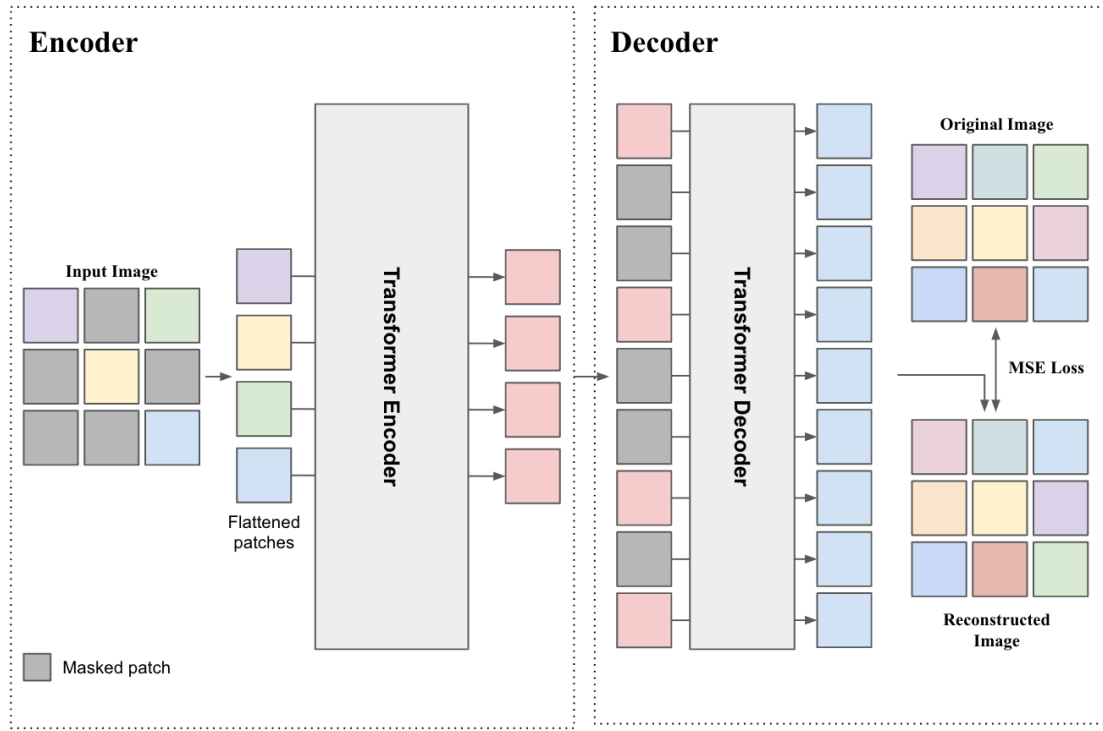

Figure 2. The Prithvi model and the pre-training architecture.

Architecture-wise Prithvi is a specialized model tailored for geospatial applications, building upon Vision Transformer (ViT) for feature extraction (Dosovitskiy *et al.* 2021), as illustrated in Figure 2. The original ViT, a groundbreaking architecture for image classification, divides an input image into fixed-size patches, linearly embeds them, and processes them through a series of attention layers (Vaswani *et al.* 2017). The result is a sequence of feature vectors, each representing the corresponding input patch, transformed based on the relationships between patches and the global context of the input. In adapting to the unique requirements of geospatial data, one of Prithvi's distinguishing features is its capability to process remote sensing data in a video format. This adaptation involves transitioning the input format from the conventional image tensor notation (C, H, W) to a more complex video tensor format (C, T, H, W), where C denotes channels, T represents time steps, H, W are the height and width of the input data. Such a modification allows the model to better capture important feature presentation by analyzing not only the spatial relationships but also the temporal relationships. For downstream tasks involving static imagery, Prithvi allows for a straightforward adjustment by setting the temporal dimension (T) to 1, ensuring flexibility in handling various types of geospatial data inputs.

During the pre-training phase, Prithvi employs a Masked AutoEncoder (MAE) learning strategy (He *et al.* 2022). The approach is particularly effective for self-supervised learning scenarios, both when labeled data is scarce or expensive and when dealing with large datasets. The MAE strategy involves masking a portion of the input data and then training the model to predict these masked values, thereby fostering a robust data representation. To facilitate this reconstruction, a decoder, consisting of a series of attention layers, is added to Prithvi. This decoder takes the encoded representations and reconstructs the original data, allowing the model to learn the intricate relationships within the data. The training process aims to minimize a Mean Squared Error (MSE) loss function, which quantifies the average squared differences between the predicted and actual values, serving as a comprehensive metric for training performance.

4.2. Task-specific adaptation of Prithvi: model head and fine-tuning

Upon completing the pre-training of Prithvi, the next step is to adapt it for specific downstream tasks. This is achieved by appending a task-specific decoder (also called a model head) to Prithvi's encoder to achieve different image analysis goals. Figure 3 shows our proposed image analysis pipeline that integrates Prithvi's pretrained encoder and is customized for object detection and instance segmentation tasks.

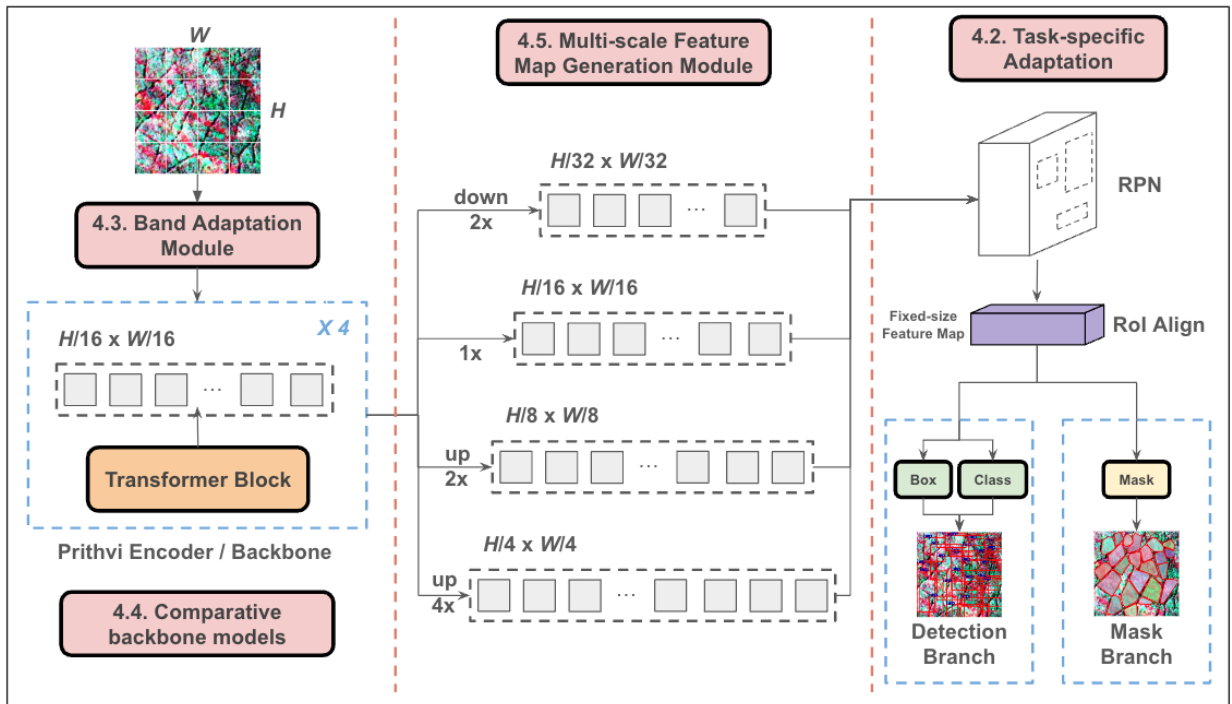

Figure 3. Image analysis pipeline for object detection and instance segmentation tasks.

To achieve these image analysis goals, we developed the pipeline utilizing the model heads (decoder) inspired by the Mask R-CNN architecture (He *et al.* 2017). As depicted in Figure 3, when using the feature map generated by Prithvi's encoder as input, the task-specific adaptation module (labeled as 4.2 in the rightmost column of the figure) begins with the Region Proposal Network (RPN), which operates on the feature map extracted from the previous stage. The RPN employs a sliding window mechanism over the feature map to generate potential bounding box proposals for objects across various scales and aspect ratios. Following the generation of these region proposals, the workflow proceeds to process the Region of Interest (RoI). The RoI Align layer is employed here to convert these proposed regions of varying sizes into fixed-sized feature maps, enabling consistent processing by downstream layers. After the RoI Align layer, each region undergoes processing by two distinct branches to generate comprehensive outputs. The first branch, the detection branch, is tasked with bounding box refinement and object category classification. It refines the initial proposals to more accurately enclose the objects and classifies each object into its respective category. Following this, the second branch, the mask branch, comes into play specifically for mask prediction for each identified object. This branch is dedicated to determining the exact pixels within the refined bounding box that constitute the object, enabling the model to produce detailed masks that delineate the object's precise shape and boundaries.

Upon this pipeline, the Prithvi model can be further fine-tuned with domain datasets for the desired tasks. When performing object detection, only the box branch is activated, whereas for instance segmentation, both the box and mask branches are employed. Training Prithvi on task-specific datasets enables the model to adapt its pre-trained knowledge to the unique characteristics of new datasets. This phase entails adjusting the weights across the entire model, including both the backbone and the appended head modules. Another strategy is to freeze the backbone weights and only train the decoder part to reduce training time and computational cost. In our study, we chose to fine-tune modules throughout the pipeline including Prithvi's backbone models so to achieve optimal performance. It is worth mentioning that this pipeline is also generalizable so the performance of Prithvi with other models can be compared by replacing the feature extractor encoder.

To further improve the adaptability and predictive performance of the Prithvi model in downstream tasks, and to enable the incorporation of the most common 3-band image as input, we enhanced the pipeline by introducing two modules: the band adaptation module and the multi-scale feature map generation module. In addition, this pipeline enables the flexible integration of other pre-trained backbone models for performance comparison. The next sections will introduce how each strategy works and conducts a comparative analysis of different models.

### 4.3. Band adaptation module

The Prithvi model is intrinsically designed to handle 6-band geospatial data, maximizing the use of the important information such multiband data provides. However, in many real-world scenarios, benchmark datasets (Bhuiyan *et al.* 2020, Schneider *et al.* 2021) may have a different band configuration than the Prithvi model. To increase the Prithvi model's applicability across diverse datasets, we developed three strategies (as shown in Figure 4) to adapt its original 6-band input to data with a different number of spectral and optical bands. The adaptation to the most commonly used 3-band RGB imagery is used as an example.

The first strategy, termed as the Zero-Padded Input, is depicted in Figure 4(b). This method involves augmenting 3-band data by appending three channels filled with zeros (depicted in black), simulating a 6-band input but devoid of any additional meaningful information. While this method seems to artificially inflate the data, it is computationally equivalent to adjusting the model's weight loading to retain only the weights associated with the existing bands. This is due to the convolutional nature of the patch embedding layer that transforms the input data. Importantly, we maintain the use of the original CNN kernels in the patch embedding layer, as demonstrated in the initially trained model shown in Figure 4(a). Although straightforward, this method might not fully tap into the model's capabilities, as it operates under the assumption that the missing three bands have a minimal bearing on the model's overall performance.

Another strategy, referred to as Channel Duplication, is detailed in Figure 4(c). In this method, the existing 3-band channels are replicated to create a 6-band input, with the duplication of red, green, and blue colors evident across the six bands. This method is based on the assumption that the initial 3 bands are sufficiently informative for the model's tasks and that the missing bands don't differ significantly in their feature-capturing capabilities compared to the available 3 bands. However, if the original 6-band model was designed to capture different types of features across all six bands, this method may not adequately substitute for that missing information. Like the Zero-Padded Input strategy, Channel Duplication also employs the same CNN kernels in the patch embedding layer, maintaining the same with the model's initial training status.

Lastly, we developed the Retrained Patch Embedding strategy, as illustrated in Figure 4(d). Rather than modifying the data to fit the model, this approach reconfigures the initial patch embedding layer of Prithvi to directly process 3-band data. Rooted in the belief that the model's architecture is versatile enough to adjust to fewer bands and that these 3 bands encompass all vital information, this method presents itself as a potentially more robust and sophisticated solution. As depicted in Figure 4(d), the CNN kernel within the patch embedding layer is reinitialized and tailored to process 3-band data, reducing the channels from 6 to 3 compared to earlier configurations. This modification not only streamlines the data processing but also reduces the model size, cutting down 590k parameters from the original model, thereby enhancing the efficiency.

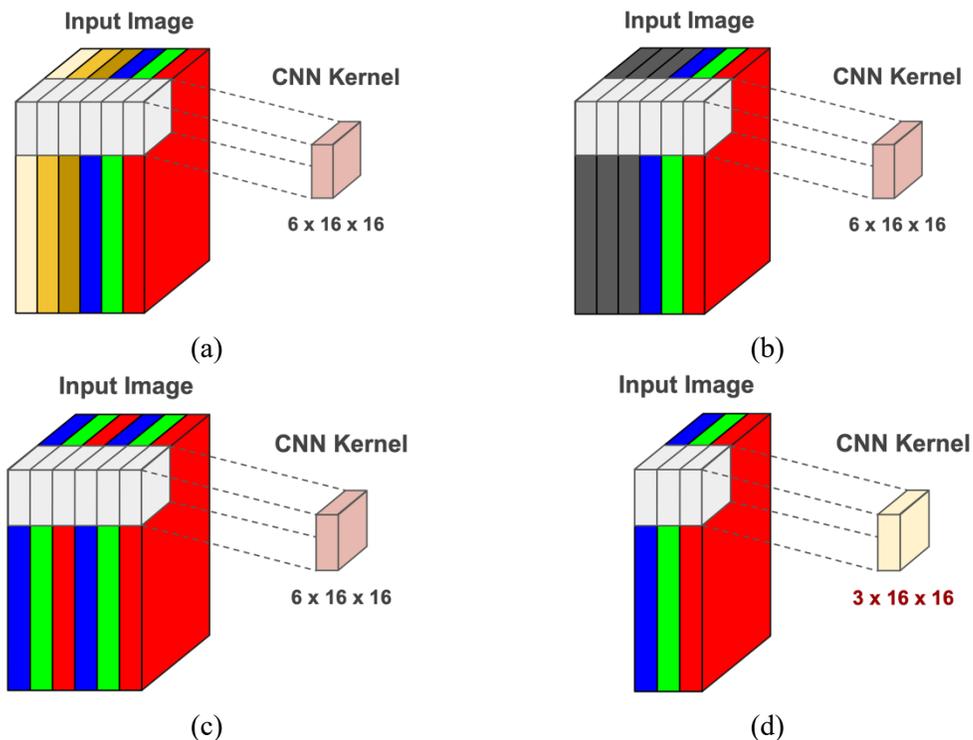

Figure 4. Input adaptation strategies. (a) Original input architecture. (b) Zero-Padded Input. (c) Channel Duplication. (d) Retrained Patch Embedding. The colors red, green, and blue represent the R, G, B band respectively.

4.4. Comparative analysis of Prithvi with other pre-trained models: benchmarking performance and adaptability in geospatial data processing

Leveraging the pipeline depicted in Figure 3, our goal in comparing Prithvi with established architecture is to pinpoint its strengths and potential areas for enhancement, with a particular focus on its training with geospatial data. To facilitate this evaluation, we have chosen three prominent, task-specific models: ViT (Li, Mao, et al. 2022), MViTv2 (Li, Wu, et al. 2022), and ResNet-50 (He et al. 2016). The comparison aims to highlight how Prithvi's unique approach to processing geospatial data compares to these well-established models. The architectures of these four models, including Prithvi, are detailed in Figure 5 (a)-(d), providing a visual reference to understand the structural differences.

ViT, as illustrated in Figure 5(a), represents a significant shift in computer vision, adopting the Transformer architecture originally developed for natural language processing (NLP). In this model, images are divided into patches, each processed akin to a token in NLP, enabling the Transformer to grasp global pixel relationships from the start. Similar to Prithvi, shown in Figure 6(c), both models operate predominantly at a single scale, producing single-scale feature maps, which aligns their approach to processing visual data. The key difference between them lies in how they pay attention to these patches. Prithvi uses a multi-head attention mechanism, which looks at the image patches in a way that considers the entire image context, akin to taking a step back to see the whole picture. On the other hand, ViT employs multi-head window attention, which means it focuses on smaller, windowed areas of the image at a time, similar to zooming in on specific details. This difference in attention methods underlines the unique ways ViT and Prithvi handle visual information, making their comparison particularly insightful for understanding how each could be best used in analyzing geospatial data.

ResNet-50, as depicted in Figure 5(b), distinguishes itself from transformer models like Prithvi and ViT by utilizing a CNN structure capable of generating hierarchical features. Unlike the single-scale focus typical of transformer architectures, ResNet-50's convolutional layers are organized hierarchically, enabling the extraction of features at multiple scales. A key aspect of ResNet-50's architecture is its residual connections, highlighted in Figure 5(b). These connections employ shortcut pathways that bypass one or more layers, directly addressing the vanishing gradient problem by allowing gradients to flow through the network more effectively. This innovation is particularly important as it enables the efficient training of deeper networks by ensuring that the added layers contribute positively to the overall performance, rather than complicating or degrading it. The operation on the residual connection is specifically designed to activate when there is a discrepancy between the input and output dimensions, ensuring smooth transitions and dimensional consistency across the network. By facilitating deeper and more efficient network architectures without loss in performance, ResNet-50 has set a benchmark in computer vision, making it an important model for comparative studies with transformer-based models.

MViTv2, illustrated in Figure 5(d), extends the Vision Transformer architecture by incorporating a unique multi-scale attention module. This key feature enables MViTv2 to emulate the multi-scale feature generation typical of CNNs such as ResNet-50, blending the extensive contextual awareness of transformers with the precise, scale-sensitive processing characteristic of CNNs. This fusion creates a hybrid model that stands out in the architectural spectrum, offering a promising avenue for comparative analysis. MViTv2's ability to produce hierarchical, multi-scale features marks it as a significant model for enhancing geospatial data analysis, bridging the divide between the singular scale focus of traditional transformer models and the layered, hierarchical structure observed in CNNs. Notably, MViTv2 incorporates a specialized pooling operation within its residual connections, a mechanism designed to activate specifically when there is a discrepancy between the input and output dimensions. This adaptive feature ensures smooth transitions across dimensions, preserving essential information without compromising the integrity of the data being processed.

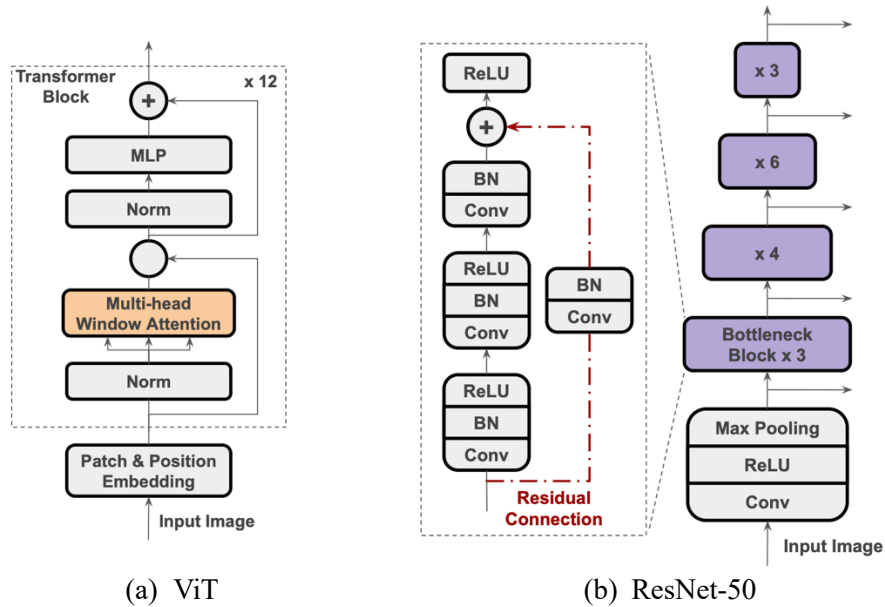

(a) ViT          (b) ResNet-50

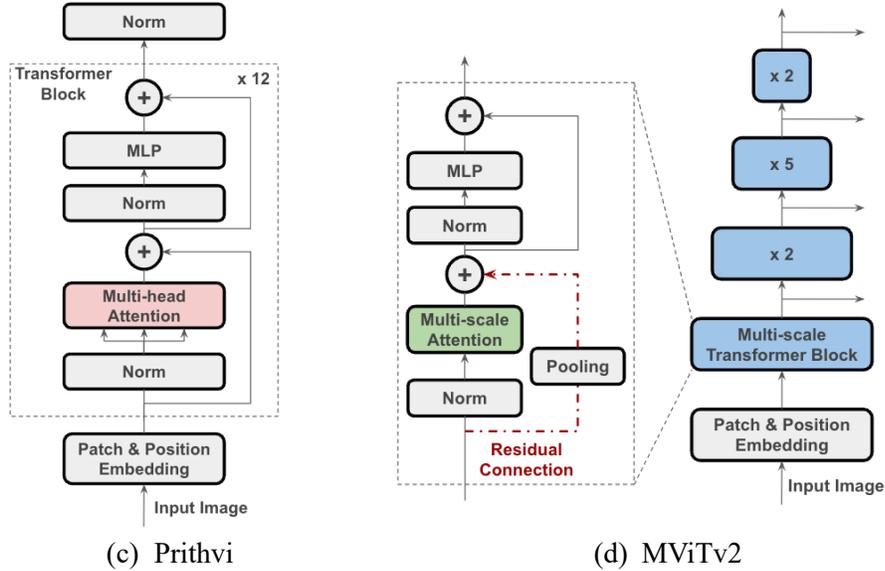

(c) Prithvi  (d) MViTv2

Figure 5. Comparable backbone model architectures. (a) ViT, (b) ResNet-50, (c) Prithvi, and (d) MViTv2

### 4.5. Multi-scale feature map generation module

In the domain of deep learning, the ability to capture multi-scale features is crucial, particularly for tasks necessitating the identification of objects or patterns across varying sizes. Multi-scale features play a pivotal role in gathering information across different scales within an image, essential for recognizing patterns across a spectrum of resolutions. This capability is particularly important in tasks such as object detection, segmentation, and recognition, where the defining characteristics of different objects may be most apparent at distinct scales. Therefore, we explore the enhancement of multi-scale features on the Prithvi model, aiming to assess how improvements in capturing these features can augment the model's performance across a range of geospatial analysis tasks.

In our model comparisons, the backbone architecture significantly influences the effectiveness of multi-scale feature extraction. We analyze two distinct types of backbones: the first type, as illustrated in Figure 6(a), is capable of generating hierarchical features, whereas the second type, shown in Figure 6(b), produces only single-scale features. Figure 6(a) represents the architecture of a Feature Pyramid Network (FPN; Lin *et al.* 2017), which generates multi-scale features within the backbone by aggregating and upsampling features across various levels of the hierarchy. For backbones limited to single-scale features, such as ViT, a multi-scale feature generation network (Li, Mao, *et al.* 2022) can be introduced, as shown in Figure 6(b), to mimic the function of an FPN. Both networks play a crucial role in generating multi-scale features, which are subsequently processed by a task adaptation head for making predictions.

In our comparative models, ResNet-50 and MViTv2 both adopt FPN shown in Figure 6(a) to generate multi-scale features. As Prithvi's backbone adopts ViT, which generates feature maps at a single scale, to enable the multi-scale feature representation capability, the strategy described in Figure 6(b) is applied to boost its performance.

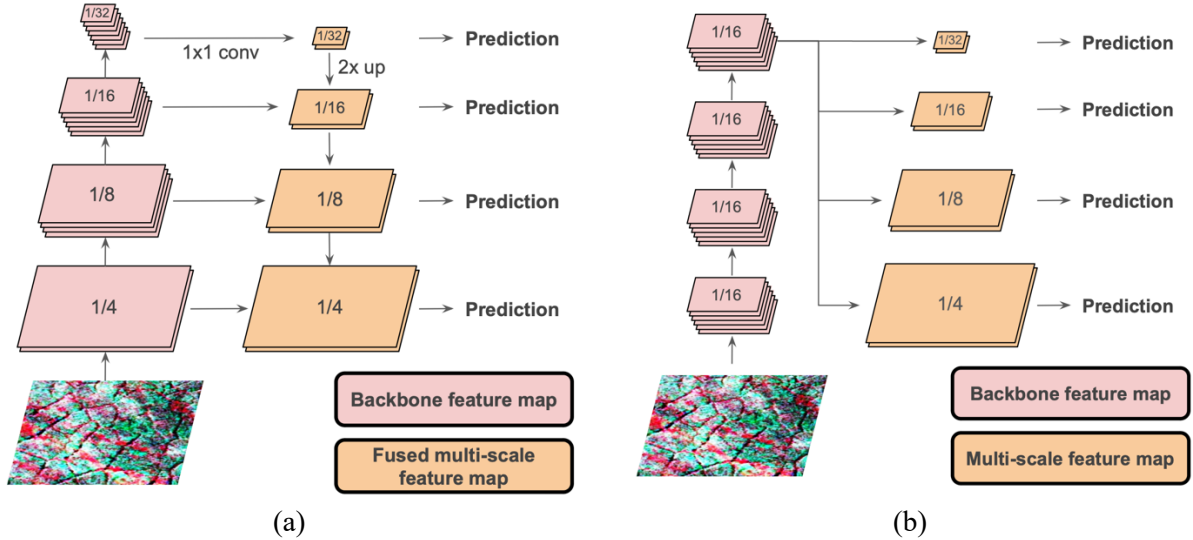

Figure 6. Multi-scale feature map generation modules. (a) A feature pyramid network for extracting hierarchical image features. (b) A multi-scale image feature generation strategy from single-scale feature maps.

## 5. Experiment

In this section, we conducted a series of experiments to assess the performance of Prithvi compared to other popular AI models. The experiments were conducted on four NVIDIA RTX A5000 GPUs, each with 24GB of memory. To evaluate model performance, mAP (Mean Average Precision) is used as the evaluation metric. This metric, widely acknowledged in the computer vision community, provides a precise measure of a model's predictive accuracy across various Intersection Over Union (IoU) thresholds, which compare the predicted area of interest (AOI) with the ground-truth AOI.

### 5.1. Input strategies: adapting 6-band Prithvi to 3-band data

While Prithvi's 6-band data is a unique aspect of the model, our study aims to assess the domain adaptability of the Prithvi model across diverse application domains, data sources, image resolutions, and geographical coverages. Given that many existing benchmark datasets may have a different band configuration than the Prithvi model, band adaptation becomes a useful feature. Additionally, because many AI models with state-of-the-art (SOTA) performance are often pre-trained with 3-band RGB data, developing an effective strategy to adapt the Prithvi input to such data or other geospatial benchmarks with a different number of input bands will expand the model's applicability. This approach will also help us compare whether there is a performance advantage of the Prithvi model pre-trained on remote sensing images over general-purpose AI models trained primarily on optical RGB images.

Therefore, to assess the adaptability of the Prithvi model, our experiment aims to evaluate the effectiveness of different input adaptation strategies, detailed in Section 4.3. To achieve this, we utilized the proposed image analysis pipeline depicted in Figure 3, integrated with Prithvi's pre-trained encoder. It was then further fine-tuned by each of the four benchmark datasets. The results presented in Table 2 highlight the comparative effectiveness of various input strategies applied across four datasets using mAP50 as the performance metric (where 50 means the IoU threshold is 50%). Among these, the Retrained Patch Embedding strategy emerged as the most effective, securing the highest mAP50 scores (0.840, 0.499, 0.483, and 0.595, respectively) and simultaneously reducing the model's size, as discussed in Section 4.3. This strategy's success is attributable to several key factors.

Table 2. Effectiveness of different band adaptation strategies on multiple datasets and tasks. Performance metric: mAP50.

| Input Strategy | Mars crater | Earth's Natural Feature | Ice-wedge polygon | EuroCrops* |
|---|---|---|---|---|
| Zero-Padded Input | 0.811 | 0.477 | 0.461 | 0.567 |
| Channel Duplication | 0.827 | 0.495 | 0.478 | 0.571 |
| Retrained Patch Embedding | **0.840** | **0.499** | **0.483** | **0.595** |

*The R, G, B bands of the EuroCrops dataset was used to test the input adaptation strategy

First, the Retrained Patch Embedding approach modifies Prithvi's architecture at a low-level, adjusting the initial patch embedding layer to efficiently handle 3-band data. This modification enables the model to leverage the full spectrum of information present in the data, eliminating the need for artificial data augmentation or manipulation. In contrast, the Zero-Padded method simply expands the dataset by appending channels of zeros, which add no real value and may distract the model from focusing on pertinent features. Similarly, the Channel Duplication method, although it maintains the integrity of the original data, only duplicates existing information. This could restrict the model's capacity to detect subtle differences within the data, owing to the resultant information redundancy.

These findings emphasize the critical role of selecting an appropriate input strategy to enhance geospatial data analysis in sophisticated models like Prithvi. By precisely aligning with both the model's architecture and the inherent characteristics of the data, the Retrained Patch Embedding strategy demonstrates its ability to significantly improve performance for adopting the popular 3-band image data as the model input.

5.2 Prithvi performance enhancement through pretrained multi-scale feature module integration

From experiments 5.1 and 5.2, the importance of multi-scale features in enhancing Prithvi's performance became evident. In this experiment, we aim to further improve upon this aspect by exploring the potential for additional refinement of Prithvi's capabilities. Our approach focuses on the integration of pre-trained weights from other models to enhance Prithvi's ability to process features more effectively.

This experiment is structured around three distinct model configurations to evaluate the impact of a multi-scale feature module on Prithvi. The first model, "Prithvi Single-Scale," employs the baseline Prithvi model, focusing exclusively on single-scale features, as investigated in experiment 5.1. The second model, "Prithvi Enhanced Multi-Scale," extends Prithvi by integrating a multi-scale feature module. This module is distinctively initialized randomly and then directly trained on downstream datasets, foregoing any pre-training, as detailed in experiment 5.2. The third model, "Prithvi Advanced Multi-Scale (Pretrained)," further evolves this approach by not only adding a multi-scale feature module to Prithvi but also enriching it with pre-trained weights from the study by Li, Mao, *et al.* (2022). This strategic integration of pre-trained weights is intended to capitalize on the extensive insights gained from training on large-scale datasets, thereby enhancing Prithvi's feature processing capabilities further.

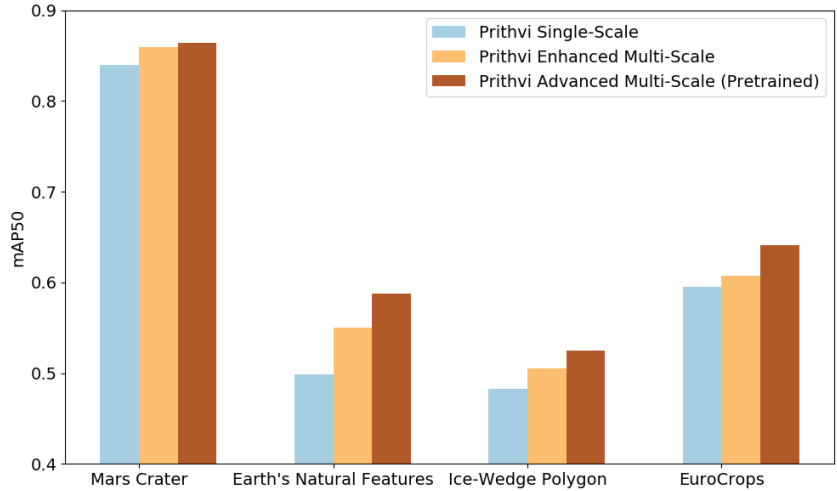

Figure 7. Prithvi model performance across datasets.

Our experimental findings, illustrated in Figure 7, highlight significant performance improvements in the Prithvi model achieved through the integration of multi-scale feature processing and the strategic application of pre-trained weights. Transitioning from its original setup, which was limited to single-scale feature processing, to advanced configurations that embrace multi-scale capabilities, Prithvi exhibits a performance boost across all four datasets. This improvement is largely credited to the inclusion of pre-trained weights, spotlighting the importance of drawing on extensive training to strengthen feature recognition across varying scales. As illustrated in Figure 7, the performance improvement is more substantial when the original model exhibits lower predictive performance (as seen in the results of the other three datasets compared to the Mars crater dataset). Performance improvement is also greater when the features are more commonly seen in other pre-training datasets, such as Earth's natural features and the EuroCrops dataset, as opposed to ice wedge polygon, which is a less commonly seen landscape feature in general AI benchmark datasets. Overall, this strategy demonstrates the potential of leveraging other models' pre-trained weights and accumulated knowledge to improve performance. It serves as a blueprint for enhancing deep learning models, suggesting that integrating pre-existing, well-trained components into new models can boost their efficiency and effectiveness.

5.3 Comparative evaluation of Prithvi with established architectures across downstream visual recognition tasks

In this experiment, we evaluate Prithvi's performance in object detection and instance segmentation, comparing it against three other models as introduced in Section 4.4. To improve their performance, all four models, including Prithvi, were enhanced with a multi-scale feature map generation module, as discussed in Section 4.5, to integrate multi-scale features effectively. Specifically, MViTv2 and ResNet-50 utilize the FPN to enrich their inherent hierarchical feature maps, as shown in Figure 6(a). Conversely, Prithvi and ViT adopt a multi-scale feature generation network, depicted in Figure 6(b), tailored to their architecture. Despite their architectural differences, each model utilizes our proposed image analysis pipeline (Figure 3) for both detection and segmentation tasks. This provides a uniform basis for comparison, as detailed in Section 4.2. To facilitate a comprehensive evaluation, all models were fine-tuned on specific downstream datasets with no weights frozen, enabling a full exploration of each architecture's capabilities.

A notable distinction lies in their initial training phases; all models, except Prithvi, were pre-trained on the ImageNet for the model backbone and COCO (The Microsoft Common Objects in Context; Lin *et al.* 2014) for the task-specific head. To ensure a fair comparison, all the models featured a pre-trained backbone combined with a newly introduced multi-scale feature map generation module and a task-adaptation head for this phase of the experiment. By implementing this strategy, we ensured that all models were evaluated under similar conditions, facilitating an accurate assessment of their performance.

In our evaluation, detailed in Table 3, Prithvi demonstrated superior accuracy, particularly in detecting Earth's natural features, with a notable mAP50 score of 0.550. This achievement not only significantly outperformed the traditional CNN model, ResNet-50, but also surpassed the scores of other transformer-based models. Prithvi's consistent performance across a variety of datasets—achieving mAP50 scores of 0.859 and 0.505 for Mars crater and ice-wedge polygons, respectively, and a score of 0.607 for the EuroCrops dataset. Prithvi also outperforms the other models on all four testing datasets using the averaged mAP score mAP[.50:.05:.95] over multiple IoU thresholds. Based on the dataset description detailed in Section 3, it is clear that the object sizes of the datasets vary. Hence, we further applied mAP measures across scales to more comprehensively understand the model's performance. As the results in Table 3 show, Earth's natural features often have large sizes (relative to the image scene), while the crop lands in EuroCrops range from relatively small to medium sizes. While there are some variations in the model's performance for segmenting objects with varying sizes, Prithvi still shows advantages over the other models, especially in mAP_S and mAP_L. This result further verifies Prithvi's adaptability and effectiveness in addressing diverse environmental and geospatial problems.

Table 3. Comparative results of Prithvi and popular supervised models on geospatial datasets. Multiple performance metrics are used here: mAP50 is the mAP with an IoU threshold of 50%); mAP [.50:.05:.95] refers to the mean average mAP over different IoU thresholds [0.5, 0.55, 0.6, 0.65, 0.7, 0.75, 0.8, 0.85, 0.9, 0.95], mAP_S, mAP_M, and mAP_L measure the mAP across different scales. Small objects are smaller than or equal to $32^2$ pixels, medium objects are between $32^2$ and $96^2$ pixels (not including $32^2$), and large objects are larger than $96^2$ pixels.

| Performance metric | Model | Mars crater | Earth's natural feature | Ice-wedge polygon | EuroCrops |
|---|---|---|---|---|---|
| mAP50 | Prithvi | **0.859** | **0.550** | **0.505** | **0.607** |
|  | ViT | 0.844 | 0.522 | 0.492 | 0.601 |
|  | MViTv2 | 0.847 | 0.540 | 0.502 | 0.594 |
|  | ResNet50 | 0.793 | 0.515 | 0.486 | 0.561 |
| mAP[.50:.05:.95] | Prithvi | **0.424** | **0.292** | **0.270** | **0.376** |
|  | ViT | 0.412 | 0.265 | 0.263 | 0.371 |
|  | MViTv2 | 0.413 | 0.281 | 0.266 | 0.364 |
|  | ResNet50 | 0.381 | 0.273 | 0.260 | 0.335 |
| mAP_S S: small | Prithvi | **0.389** | N/A | **0.246** | **0.367** |
|  | ViT | 0.374 | N/A | 0.238 | 0.365 |
|  | MViTv2 | 0.377 | N/A | 0.243 | 0.360 |
|  | ResNet50 | 0.346 | N/A | 0.235 | 0.331 |
| mAP_M M: medium | Prithvi | 0.639 | 0.127 | **0.429** | **0.477** |
|  | ViT | 0.637 | **0.144** | 0.422 | 0.448 |
|  | MViTv2 | **0.646** | 0.111 | 0.417 | 0.435 |
|  | ResNet50 | 0.623 | 0.139 | 0.415 | 0.393 |
| mAP_L L: large | Prithvi | **0.829** | **0.303** | **0.597** | N/A |
|  | ViT | 0.794 | 0.272 | 0.574 | N/A |
|  | MViTv2 | 0.806 | 0.296 | 0.596 | N/A |
|  | ResNet50 | 0.771 | 0.284 | 0.576 | N/A |

The direct comparison of Prithvi with ViT, considering their architectural similarities, offers valuable insights. Here, except for mAP_M on the Earth's natural features dataset, Prithvi consistently demonstrates better performance than the ViT architecture. This comparison highlights the benefits of pre-training on extensive remote sensing datasets, which likely contributed to Prithvi's better performance over ViT. Furthermore, the comparison between ViT and MViTv2 (with stronger performance observed for MViTv2 than for ViT) highlights the importance of incorporating hierarchical features early in the pre-training process. MViTv2's ability to seamlessly integrate these features showcases a flexible approach to visual data processing that leverages the strengths of both transformer and CNN architectures.

Besides the prediction accuracy, we also compared the computational efficiency of different model architectures in terms of both training and inference speed. Table 4 lists the results. The numbers in the table are averaged speed over the four datasets. In fact, because all the input images, regardless of their sizes, will be transformed by all models into a fixed-size image before going through the image analysis pipeline, the time cost is independent of input datasets.

Table 4. Computational efficiency of the Prithvi model

| Model | Inference time (s) per image | Training time (s) per iteration |
| --- | --- | --- |
| Prithvi | 0.297 | 0.657 |
| ViT | 0.263 | 0.441 |
| MViTv2 | 0.164 | 0.319 |
| ResNet50 | 0.097 | 0.265 |

Table 4 results show that Prithvi and ViT exhibit similar inference and training speeds due to their analogous architectures, as shown in Figure 5. ViT's multi-head window attention, compared to Prithvi's multi-head attention, gives it a slight speed advantage. Both Prithvi and ViT are slower than MViTv2 and ResNet50, whose hierarchical structures progressively reduce the size of processed feature maps, leading to faster computations. ResNet50's particularly faster speed than the other transformer-based models is attributed to its efficient residual connections and simpler architecture. In summary, while Prithvi has shown a clear advantage in prediction accuracy, it runs slower than the other comparative models. Hence, it is more suitable for applications requiring high result accuracy and demanding less in computational efficiency.

5.4 Impact of image resolution on model performance
This experiment further explores the impact of input image resolution on model performance. We selected two datasets, ice-wedge polygons and EuroCrops, and the two best-performing models, Prithvi and MViTv2, for the comparison. For each dataset, we reduced the image resolution to 1/2, 1/4, and 1/8 of the original value and tested the model performance. The results, as shown in Table 5, indicate that both Prithvi and MViTv2 experience reduced performance as image resolution becomes coarser, but the trend of performance decrease is similar, even though the two models are pretrained on different images (HLS vs. ImageNet). For example, on the EuroCrops dataset, when the image resolution is reduced to half (from 10m to 20m), the performance of both the Prithvi and MViTv2 models decreases substantially (29.16% for Prithvi and 27.78% for MViTv2). When the resolution is reduced further, a more significant decrease is observed (Table 5). In contrast, when the image resolution for the ice-wedge polygon dataset is reduced to half of its original size (from 0.5m to 1m), there is only a slight reduction in model performance. When the resolution of the input image is reduced further, the performance of both Prithvi and MViTv2 drops significantly.

This difference in results is less likely due to the discrepancy in image resolution between the pre-training images and the testing images of the foundation model, but rather due to the relative object size and the level of detail captured in the image scene. For EuroCrops, which uses Sentinel-2 data similar to the Prithvi pre-training data, when the resolution of the testing image decreases, we did not observe a performance advantage for Prithvi compared to MViTv2, which is pretrained on ImageNet. This is because the cropland boundaries in EuroCrops are not very sharp in the original image, so decreasing the image resolution makes the object boundaries blurrier and therefore more difficult to detect. However, even though the absolute size of ice-wedge polygon is small, because super high-resolution imagery (0.5m) is used for the analysis, it maintains a good size in the image. Thus, even when reducing the resolution to half, the features are still large and clear enough to be detected. Hence, our conclusion is that while input image resolution does affect model performance, the impact is more dataset-dependent and less dependent on the consistency of image resolutions between the pre-training and testing images.

Table 5. Model performance (measured by mAP50) with different input image resolutions.

| | Model | 10m (Original) | 20m | 40m | 80m |
|---|---|---|---|---|---|
| EuroCrops | Prithvi | 0.607 | 0.430 (-29.16%) | 0.251 (-58.65%) | 0.154 (-74.63%) |
| | MViTv2 | 0.594 | 0.429 (-27.78%) | 0.259 (-56.40%) | 0.141 (-76.26%) |
| Ice-wedge polygons | | .5m (Original) | 1m | 2m | 4m |
| | Prithvi | 0.505 | 0.503 (-0.4%) | 0.399 (-20.99%) | 0.166 (-67.13%) |
| | MViTv2 | 0.502 | 0.496 (-1.2%) | 0.388 (-22.71%) | 0.171 (-65.94%) |

5.6 Data effectiveness of Prithvi in enabling few-shot learning

One desired property of a foundation model is the enablement of zero-shot or few-shot learning, and this capability is well-documented in large language models (LLMs). In this section, we conducted an ablation study using 75%, 50%, and 25% of the training data to assess Prithvi's few(er)-shot learning capability. As shown in Table 6, Prithvi's performance has been quite stable when reducing the training datasets from 100% to 75%, with no or very slight reduction in predictive performance. When reducing the training dataset to 50%, a more noticeable decrease is found for processing Earth's natural features and EuroCrops datasets. The model performance on Mars crater and ice-wedge polygon datasets remains quite good. This is likely because the features in Mars craters and ice-wedge polygons are more similar to each other, so with fewer samples, the model is still good at making predictions. For the other two datasets, which have more diversity in their training samples and feature types, more performance variance is observed. Overall, the Prithvi model shows very good data efficiency, with less than a 10% performance drop observed for the Mars crater, ice-wedge polygon, and EuroCrops datasets when reducing the training samples from 100% to 25%. The models show a more significant reduction (19%) in predictive performance for the Earth's natural features, likely due to the smaller total number of samples in the dataset compared to the others (refer to Table 1 for dataset characteristics).

Table 6. Prithvi's capability in enabling few-shot learning. The predictive accuracy values reported are mAP50.

| Percentage of training data used | Mars crater | Earth's natural feature | Ice-wedge polygon | EuroCrops |
|---|---|---|---|---|
| 100% | 0.859 | 0.550 | 0.505 | 0.607 |
| 75% | 0.858 | 0.550 | 0.502 | 0.602 |
| 50% | 0.855 | 0.520 | 0.503 | 0.586 |
| 25% | 0.848 | 0.444 | 0.458 | 0.580 |

6. Discussions on the strengths and limitations of Prithvi

In previous experiments, we explored the effectiveness of the band adaptation strategy and multi-scale feature for adapting and enhancing Prithvi for diverse geospatial tasks. Since Prithvi contains only a backbone model pre-trained on multi-spectral datasets, it lacks a fully released pipeline trained across all its components: backbone, multi-scale feature generator, and detection/segmentation head. In contrast, some other models, such as MViTv2, are already pre-trained on large AI benchmark datasets across all these modules to achieve optimal performance during model adaptation. To ensure a fair comparison, in the experiments from Section 5.1 to Section 5.3, we maintained consistent experimental conditions across all models, including the use of the multi-scale module and the use of pre-trained weights in the detection/segmentation head. These experiments demonstrated the power of Prithvi's backbone model in adapting to remote sensing image analysis tasks due to the knowledge learned from similar data during pre-training.

In this section, we conduct further experiments to identify areas for improvement in Prithvi. We compared the Prithvi model with the MViTv2-Optimal model, which is fully pre-trained and optimized on the entire segmentation pipeline with multi-spectral data input. The EuroCrop dataset is used as the experimental dataset because it contains the 6-band data required for Prithvi. Since the pre-trained MViTv2-Optimal can only take 3-band data as input, only the RGB band of the EuroCrop data is sent to MViTv2-Optimal.

Table 7. A comparison between Prithvi and the optimal configuration of MViTv2 on 6-band data

| Model | EuroCrops (6-band) | EuroCrops (3-band) |
| --- | --- | --- |
| Prithvi | 0.657 | 0.641 |
| MViTv2-Optimal | N/A | 0.708 |

Table 7 shows the comparative results. When Prithvi takes 6-band input, the model's predictive performance improves (from 0.641 to 0.657) compared to when it is given only 3-band input. This improvement validates that incorporating a broader spectral range can enhance a model's capacity to interpret and analyze remote sensing imagery, underscoring the value of accessing extensive spectral information to boost the effectiveness of deep learning models.

When the Prithvi model taking 6-band input is compared with the optimized MViTv2 taking 3-band input, it still shows a performance gap. This result emphasizes the value of conducting full pre-training/fine-tuning across task-specific pipeline (as the MViTv2-Optimal did). And this feature is critical to achieve a SOTA performance to fully demonstrate the value of multi-spectral remote sensing training data in relevant analysis. In fact, in other foundation model such as Microsoft's Florence model (Yuan *et al.* 2021), the SOTA performance is achieved not only because of the backbone model pretrained on a massive amount data (which is highly important), the model is further carefully trained on large datasets supporting downstream tasks. Despite the power demonstrated by Prithvi's backbone model, pre-training on the entire pipeline is a clear area for improvement for the Prithvi model.

It is also worth mentioning that a fully trained task-specific pipeline is critical to improving Prithvi's data efficiency and enabling few-shot learning. In Section 5.6, we demonstrated the stable performance of Prithvi when substantially reducing the training sample size in downstream tasks. However, there are still at least 100 samples used in each benchmark dataset to fine-tune the Prithvi model due to the need to train new heads and adaptation layers. When a well-trained task-specific pipeline becomes available, there is an opportunity for Prithvi (and other vision foundation models) to enable domain adaptation with only a few data samples.

7. Conclusion and Future Work

This paper evaluates the effectiveness of NASA-IBM's foundation model Prithvi in its ability for multiple downstream tasks for remote sensing image analysis. Four benchmark datasets containing environmental and land use features and covering diverse geographical regions are selected in the analysis. Through a series of experiments, we demonstrated the advantages of the Prithvi model in gaining useful geospatial knowledge from multi-spectral HLS data and its effectiveness in object detection and segmentation tasks compared to other large task-specific AI models. Besides evaluation, we have also proposed and developed an image analysis pipeline that can incorporate multiple backbone models with enhancement strategies to further boost up Prithvi's performance. The patch embedding strategy improves the Prithvi model's data adaptability, and the multi-scale feature generation further enhances Prithvi's feature extraction capability.

However, we also identified a weakness in Prithvi, specifically its lack of a fully trained task-specific pipeline despite its powerful backbone model for geospatial analysis. The analysis in Section 6 demonstrates that although the incorporation of multi-spectral data is crucial in remote sensing image analysis, its effectiveness may not be well showcased without an optimized model pipeline carefully trained with large datasets. Despite fine-tuning the proposed pipeline (Figure 3) with smaller, domain-specific datasets, the introduced parameters may not be well trained due to the limited scale of relevant data. Therefore, a potential area for Prithvi's improvement is further fine-tuning major image analysis pipelines with its pre-trained backbones to enhance performance in downstream tasks. Achieving this, however, is non-trivial, demanding substantial computing power and the exploration of new techniques for model training and behavior monitoring. A community-driven approach and collaboration among academia, industry, and government agencies may provide valuable insights to collectively advance in this direction.

As geospatial foundation model research has been rapidly advancing, it is very important to develop a standardized benchmarking approach to ensure the thorough evaluation and fair comparison of existing models. In preparing the benchmarking data, it is crucial to prevent data leakage, a known issue where general foundation models memorize content from pre-training data (Chen, Li, *et al.* 2024, Xu *et al.* 2024). For the evaluation of geospatial foundation models, especially vision models, four strategies could help mitigate this issue: (1) **Geographical diversity**: Since geospatial data are often associated with specific locations, ensuring that benchmarking datasets come from a wide range of geographical locations can prevent overlap between pre-training and testing data. In our case, multiple datasets such as Mars crater, Arctic Ice Wedge Polygon, and EuroCrops have different geographical coverages (Martian surface, Arctic, and central Denmark) from Prithvi's pre-training data centered on the continental US. (2) **Transparency**: Clear documentation of benchmark data utilization should be strongly encouraged in the GeoAI research community. Filling out a "benchmark transparency card" (Xu *et al.* 2024) with a list of datasets and evaluation-related questions will help improve transparency and clarity, facilitating the healthy development of geospatial foundation models. (3) **Multimodality**: Geospatial data is very rich in terms of data source, resolution, and spectral bands. Leveraging data acquired from diverse satellite platforms will help evaluate transferability in Earth science applications across data modalities. In Prithvi's case, we adopted datasets from diverse input sources and application areas to achieve a comprehensive view of the model's transferability and adaptability. (4) **Spatial autocorrelation**: A significant difference between GeoAI and general AI research is its location relevance. Because of this, GeoAI datasets will possess the property of being spatially correlated, indicating that geospatial data from nearby locations are more similar to each other. To prevent data leakage issues, it is important to identify potential spatial autocorrelation within the benchmark data to avoid oversampling pre-training data that have more geographical proximity to the testing data. We hope these discussions will facilitate further dialogue toward developing a sustainable ecosystem for geospatial foundation model and GeoAI research.

In conclusion, this research contributes to a more comprehensive understanding of geospatial foundation models by exploring the benefits of pre-training with remote sensing imagery and important aspects leading to stronger predictive performance in real-world applications. We have also demonstrated methods to adapt and enhance the data and application adaptability of the Prithvi model. The knowledge gained from this study is intended to be valuable for geospatial researchers interested in integrating geospatial foundation models (GFM), like Prithvi, into their research. Our work also provides insights into the construction and fine-tuning of future GFM for achieving optimal performance.

**Acknowledgements**

This work is supported in part by the National Science Foundation under awards 1853864, 2120943, 2230034, as well as Google.org's Impact Challenge for Climate Innovation Program.